\documentclass[10pt,twocolumn,letterpaper]{article}
\usepackage{iccv}

\usepackage{graphicx}
\usepackage{amsmath}
\usepackage{amssymb}
\usepackage{booktabs}
\usepackage{multirow}
\usepackage{subfigure}

\usepackage{color, colortbl}
\definecolor{Gray}{gray}{0.9}
\usepackage{xcolor}

\usepackage[pagebackref=true,breaklinks=true,letterpaper=true,colorlinks,bookmarks=false]{hyperref}

\usepackage[capitalize]{cleveref}
\crefname{section}{Sec.}{Secs.}
\Crefname{section}{Section}{Sections}
\Crefname{table}{Table}{Tables}
\crefname{table}{Tab.}{Tabs.}

\usepackage{algorithm}
\usepackage{algpseudocode}

\usepackage{mathtools}

\iccvfinalcopy % *** Uncomment this line for the final submission

\def\iccvPaperID{7047} % *** Enter the ICCV Paper ID here

\ificcvfinal\fi

\begin{document}

\title{Inductive Attention for Video Action Anticipation}

\author{
Tsung-Ming Tai$^{1,2}$, Giuseppe Fiameni$^{1}$, Cheng-Kuang Lee$^{1}$, Simon See$^{1}$, Oswald Lanz$^{2}$\\
$^1$NVIDIA AI Technology Center \quad $^2$Free University of Bozen-Bolzano\\
{\tt\small \{ntai, gfiameni, ckl, ssee\}@nvidia.com \quad \{tstai, oswald.lanz\}@unibz.it}}

% \and
\maketitle

%%%%%%%%% ABSTRACT
\begin{abstract}
Anticipating future actions based on spatiotemporal observations is essential in video understanding and predictive computer vision. Moreover, a model capable of anticipating the future has important applications, it can benefit precautionary systems to react before an event occurs. However, unlike in the action recognition task, future information is inaccessible at observation time -- a model cannot directly map the video frames to the target action to solve the anticipation task. Instead, the temporal inference is required to associate the relevant evidence with possible future actions. Consequently, existing solutions based on the action recognition models are only suboptimal. Recently, researchers proposed extending the observation window to capture longer pre-action profiles from past moments and leveraging attention to retrieve the subtle evidence to improve the anticipation predictions. However, existing attention designs typically use frame inputs as the query which is suboptimal, as a video frame only weakly connects to the future action. To this end, we propose an inductive attention model, dubbed IAM, which leverages the current prediction priors as the query to infer future action and can efficiently process the long video content. Furthermore, our method considers the uncertainty of the future via the many-to-many association in the attention design. As a result, IAM consistently outperforms the state-of-the-art anticipation models on multiple large-scale egocentric video datasets while using significantly fewer model parameters.
\end{abstract}

%%%%%%%%% BODY TEXT
\label{sec:intro}
\begin{figure}[t]
    \centering
    \includegraphics[width=1\columnwidth]{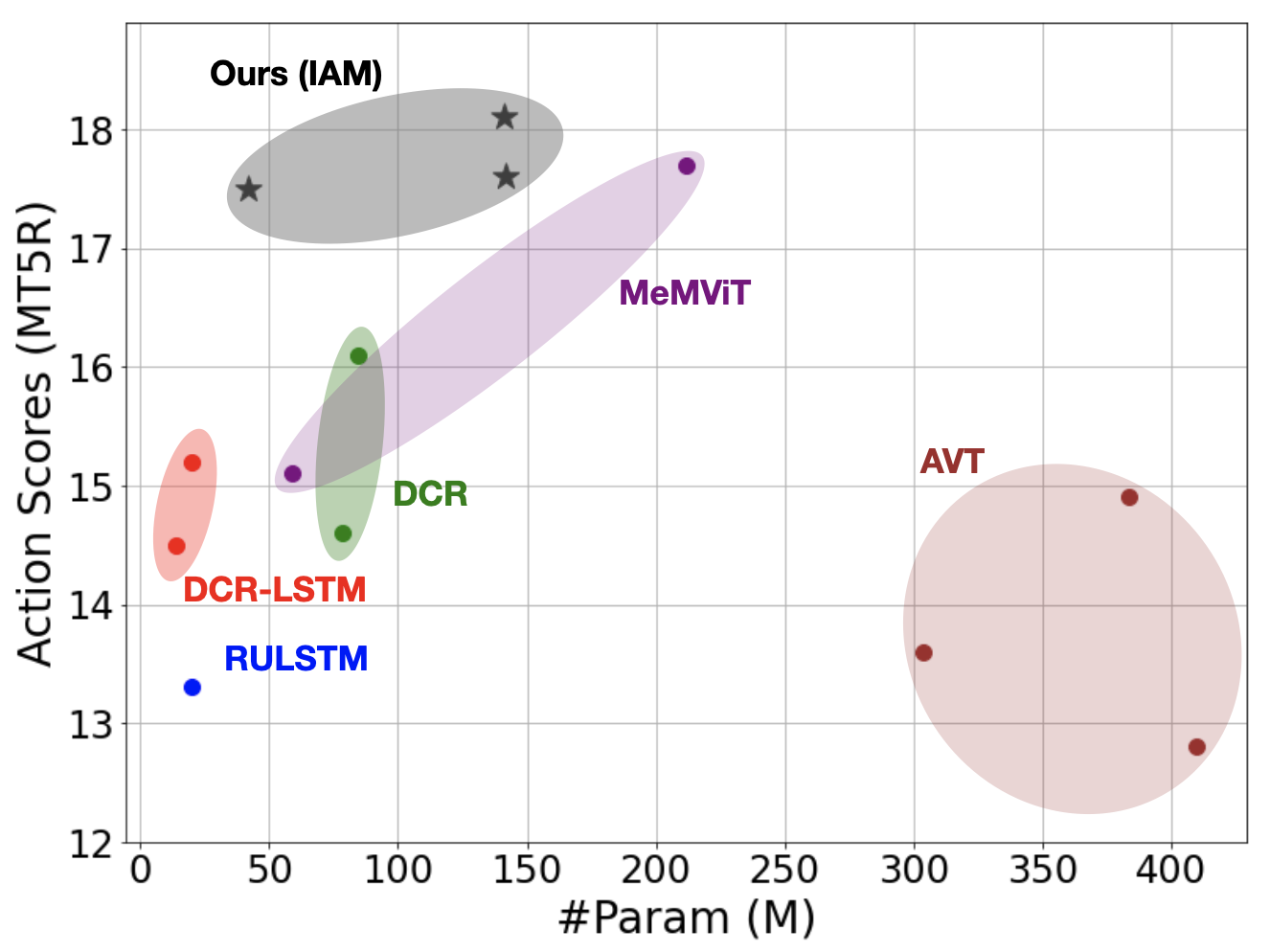}
    \caption{{\bf Inductive Attention Model} (IAM) achieves higher action anticipation accuracy with fewer parameters. Here we compare IAM to previous methods on the EPIC-Kitchens-100 dataset. (Multiple points indicate model variants using different backbones.)}
    \label{fig:param_plot}
\end{figure}

\section{Introduction}
Recognizing human actions from videos has been extensively studied in video understanding \cite{DBLP:conf/icml/BertasiusWT21,DBLP:conf/cvpr/GirdharCDZ19,DBLP:conf/cvpr/0004GGH18,DBLP:conf/nips/ChenKLYF18,DBLP:journals/cviu/LiGGJS18,DBLP:conf/nips/GirdharR17,DBLP:conf/cvpr/SudhakaranEL19,DBLP:conf/iccv/LinGH19,DBLP:conf/cvpr/SudhakaranEL20,sudhakaran2021learning}. Since the input observations are aligned in time with their action annotations, the action recognition models can learn the direct mapping of the video context to the action semantics. Besides, unlike action recognition, early action recognition foresees the action with incomplete observations of input videos; the models only rely on initial frames to describe an action \cite{su2020convolutional,wang2018eidetic,DBLP:conf/wacv/GeestT18,DBLP:conf/cvpr/KongTF17,DBLP:conf/iccv/AkbarianSSFPA17,DBLP:conf/cvpr/MaSS16}. Although early action recognition models are capable of detecting an activity as early as possible, the partial observations of action is still required as inputs. Therefore for some applications, such as precautionary systems \cite{anderson2010preemption, ricci2003precaution, reid2022anticipating} that must plan before an event occurs, %solving with the 
early action recognition could not satisfy the requirements.

Video action anticipation, on the other hand, aims to predict near-future actions based on disjoint pre-action intervals \cite{gollwitzer1990action, moltisanti2017trespassing}. In this case, action is completely hidden from the supporting video input, leading to different challenges than in the action recognition or early action recognition tasks. For example, observations can no longer describe an action but only hint to 
%prophesies 
the occurrence of specific further action. Furthermore, similar pre-actions could lead to different possible consequence actions, where the uncertainty of the future reveals a many-to-many association rather than a one-to-one mapping. As a result, action anticipation is more akin to causal inference than to pattern recognition. Addressing the anticipation task with existing recognition models is suboptimal.

The advent of transformers contributed to the achievement of remarkable performance in action anticipation \cite{girdhar2021anticipative, wu2022memvit}. Recent work \cite{wu2022memvit} proposes using a more extended context to compensate for insufficient information on pre-action intervals and shows improving accuracy in action anticipation. However, the underlying design remained unchanged from that of action recognition. One of the critical design choices in attention models is the selection of an appropriate query that triggers the discovery of valuable features. A widely adopted option is to take the incoming frame as the input query in sequential order. However, %guiding 
querying %through perception input 
by observing may be less effective and an indirect usage, considering the weak connection between the video content and anticipation labels.

%{\color{blue}
An alternative, and in this context arguably more grounded choice for the query is to leverage the last prediction directly. In object detection and tracking \cite{carion2020end,zhudeformable} a lookahead predictor generates the future estimate based on the last predictions fused with the supporting observation. Likewise, in action anticipation a lookahead predictor can be employed to generate a prediction in the action space.

To this end, we propose an inductive attention model, dubbed IAM, %for video action anticipation, 
which is capable of processing the long video inputs and uses the previous anticipative action prediction as a query to induce possible future actions from experience. We equip IAM with an indexable memory
to store past contexts for attention lookup, where each past moment is recorded in a key-value pair formed by representative features and corresponding anticipation scores. Our proposed model natively considers many-to-many associations while computing the correlation between the last prediction and previously predicted action trajectories (stored as the memory keys). As a result, the relevant pre-action durations can be retrieved and associated with anticipating future action. %As the experiment section demonstrates, our proposed 
As shown in Figure~\ref{fig:param_plot}, our model can significantly improve action accuracy over the previous methods. IAM obtains state-of-the-art accuracy on large-scale anticipation datasets, despite using fewer model parameters. 

Our proposed IAM recurrently updates the internal states at each video timestep. Empirical findings in previous works \cite{wang2018eidetic, su2020convolutional} showed that addressing the predictive tasks using sequential modeling enables inferring the temporal connection better than with the clip-based approach. A hypothesis behind this is that recurrent networks can efficiently capture the sequence manifold, which was associated with an inductive bias in previous works \cite{snyders2001inductive, feinman2018learning}. Consequently, transformers designed for action anticipation employ a causal decoder and consume inputs sequentially to encode the sequence dynamics \cite{girdhar2021anticipative, wu2022memvit}. Furthermore, to address the well-known forgetting problem of recurrent models \cite{hochreiter1997long, chung2014empirical}, our model leverages a higher-order recurrent design \cite{soltani2016higher, su2020convolutional, tai2022higher} in the update process, in conjunction with inductive attention and the indexed memory.

\label{sec:related}
\section{Related Work}
\noindent \textbf{RNNs in Action Anticipation.}
Recurrent neural networks (RNNs) are commonly used to model the sequence input. Some empirical evidence shows the inductive bias learned by the recurrent models can effectively handle action prediction tasks \cite{wang2018eidetic, su2020convolutional}. Early works on video action anticipation employed LSTM in an encoder-decoder architecture and established the baseline in the video action anticipation problem \cite{furnari2020rolling}. Several subsequent improvements are proposed over the baseline, \cite{osman2021slowfast} enhanced the original LSTM encoder with an encoder that can handle the spatial-temporal structure; \cite{qi2021self} introduced a self-regulated module and achieved accuracy gains by exploiting long-range context features in recurrent updates. In addition, \cite{camporese2021knowledge} studied using label smoothing to account for future uncertainty and shows improvements. On the other hand, some related works extended the recurrent mechanism, for example, \cite{wu2020learning, fernando2021anticipating} explicitly leveraging future information and simultaneously predicting future frames; \cite{tai2021higher}  leveraged higher-order recurrent and combined with proposed spatial-temporal decomposed attention for action predictions. Moreover, \cite{tai2022unified} generalized the recurrent mechanism as message-passing learning and analyzed the information propagation in spatial-temporal with self-attention. 

Our work also leverages higher-order recurrence in the updates, and incorporates inductive attention to predict future action. As a result, our proposed model can explicitly lookup the previous 30s contexts in each recurrent update to alleviate the forgetting problem in conventional RNNs.
\newline

\begin{figure}[t]
    \centering
    \includegraphics[width=1\columnwidth]{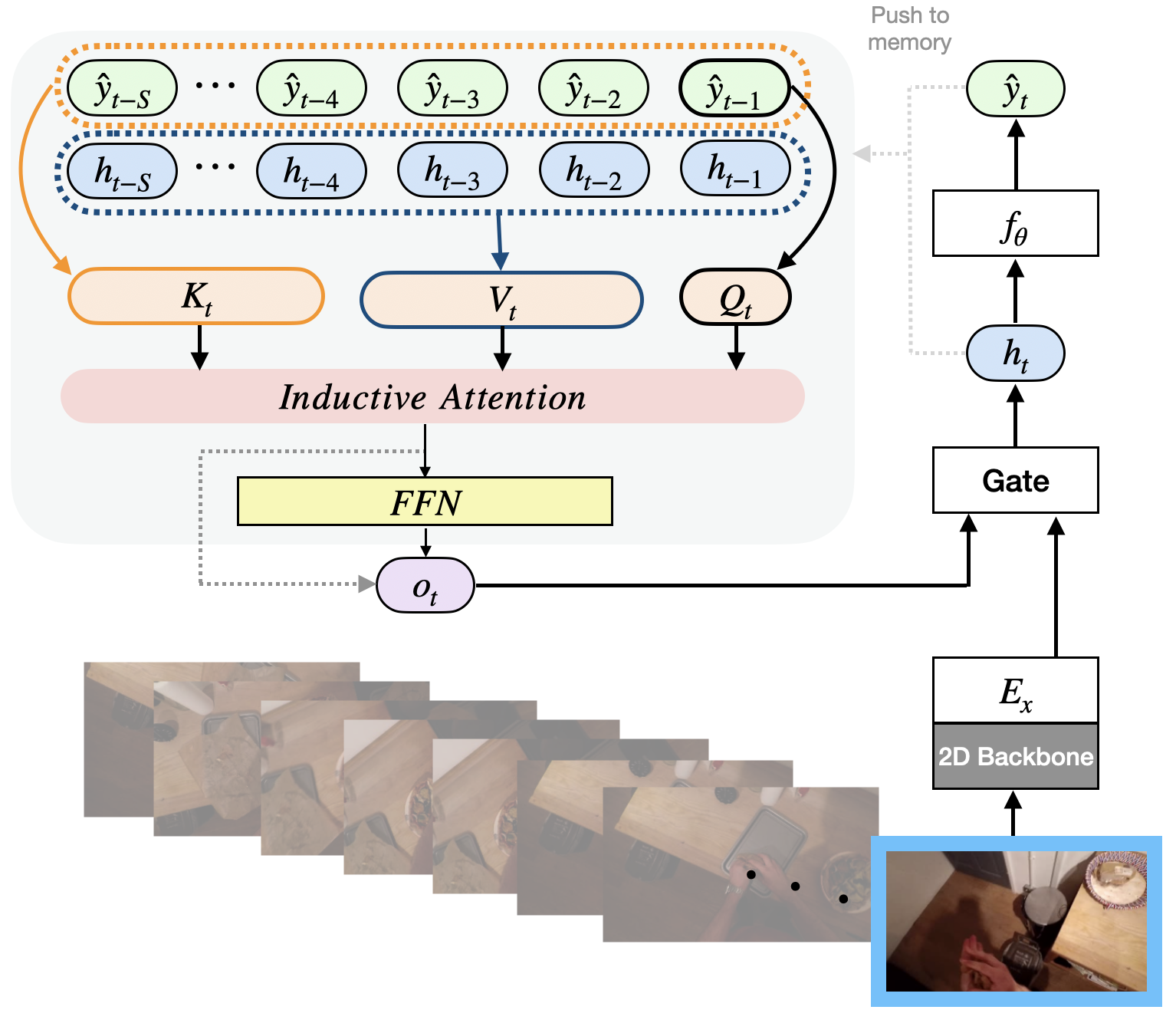}
    \caption{Inductive attention model takes the previous prediction $\hat{y}_{t-1}$ as the query and computes the correlation with memory keys ($\hat{y}_{t-1},\dots,\hat{y}_{t-S}$) to aggregate a maintained queue of recurrent states ($h_{t-1},\dots,h_{t-S}$). The output is then fused with the encoded frame input via a gating function and forms the recurrent state $h_t$ to derive the current prediction $\hat{y}_t$.}
    \label{fig:recurrent_cell}
\end{figure}

\noindent \textbf{Transformers in Action Anticipation.}
Transformer architectures have been applied with success in video action recognition \cite{bertasius2021space, arnab2021vivit}. The core element in transformers is self-attention \cite{vaswani2017attention}, originally introduced for language processing tasks. Later, vision transformers (ViTs) led the trend in various vision tasks \cite{dosovitskiy2020image, zhou2021deepvit, han2022survey}. Recent works on action anticipation build upon transformers as a basic model, outperforming recurrent baselines. \cite{girdhar2021anticipative} proposed a transformer model solely based on self-attention layers for action anticipation. The model comprises the ViT backbone and the causal decoder to capture the temporal dependencies. \cite{wu2022memvit} proposed a memory compression module for the transformer to model long-range dependencies efficiently and has nearly linear scaling complexity, providing outstanding performance in various video tasks.

Transformer has shown capability in general modeling for sequential inputs \cite{jaegle2021perceiver, jaegle2021perceiverio}. In line with these findings, we deploy attention in our model as a flexible mechanism for distilling temporal context. Our work is also related to the recent progress of combining sequence modeling and attention \cite{huangencoding,DBLP:journals/corr/abs-2203-07852,DBLP:journals/corr/abs-2207-06881}. These approaches integrate recurrent with self-attention to overcome the quadratic computation complexity of transformers and to learn better representations for recurrent propagation. However, these methods have not yet been extensively evaluated on large datasets nor tested in practical application settings.
\newline

% \noindent \textbf{Video Action Anticipation in Online Streaming.}
% Most previous work considers action anticipation performed offline, i.e., without constraints on computational resources \cite{furnari2021towards}, or anticipates only in trimmed videos at a fixed, predefined anticipation interval \cite{rodin2022untrimmed}. Although the most popular benchmark datasets are designed to test the model using trimmed videos, our proposed model can support online and untrimmed video processing. It is based on a lightweight recurrent update and can output the anticipated result anytime. We also extend the memory capacity to capture the long-range temporal dependency without increasing model parameters. 

\label{sec:method}
\section{Preliminaries}
\noindent \textbf{Higher-Order Recurrent Networks.}
Similar to conventional n-gram language model \cite{shannon1948mathematical}, higher-order recurrent networks \cite{soltani2016higher} sequentially compute new outputs and internal states from aggregations of multiple previous states. It is considered the extension of the (first order) Markov chain assumption behind conventional recurrent networks. 

More formally, in a first-order recurrent model, the new internal (hidden) state $h_t$ is computed from the input $x_t$ and a gated version of the previous hidden state, $g(h_{t-1})$, as follows 
\begin{equation}
    h_t = f(x_t + g(h_{t-1})).
\end{equation}
In a higher-order recurrent model, multiple past states are referenced and aggregated to predict the new state, by
\begin{equation}
    h_t = f(x_t + \phi(h_{t-1},\dots,h_{t-S})). \label{eq:higher-order}
\end{equation}
Here $S$ is a hyper-parameter to specify the number of past states (the order of the model), and function $\phi(\cdot)$ implements the (gated) aggregation in the higher-order model.

There are several choices for the aggregation function $\phi$, such as linear function \cite{soltani2016higher}, polynomial function \cite{yu2017long}, convolutional tensor-train decomposition \cite{su2020convolutional}, spatial-temporal decomposed attention \cite{tai2021higher}. In this work, we propose inductive attention to summarizing from the higher-order states.
\newline

\noindent \textbf{(Self-)Attention.}
Attention retrieves the correlation context in the values $V$ by computing the dot-product correlation of keys $K$ conditioned on the query $Q$. Typically, $V$ and $K$ can be traced back to the same input source, while the query $Q$ is purposely chosen to ensure task-relevant information is retrieved by attention. The standard form of attention \cite{vaswani2017attention} is implemented as follows
\begin{equation}
    Attention(Q, K, V) = softmax(\frac{Q^TK}{\sqrt{d}})V \label{eq:att}
\end{equation}
where $d$ denotes the dimension of vectors $K$ and $Q$. Self-attention refers to the case when $Q, K, V$ are all derived from the same input source. Multi-Head Attention (MHA) further separates the $Q, K, V$ into multiple groups (heads) in parallel and fuses heads with an additional linear transform at the output.

Since the attention layer contains only linear operations, a Feed-Forward Network (FFN) comes in for non-linear layer stacking. FFN is implemented by two fully-connected layers, with the expanding factor of 4 for the bottleneck, and a non-linear activator function $act$ placed in between
\begin{equation}
    FFN(x) = act(w_1^T norm(x))w_2 + x  \label{eq:ffn}
\end{equation}
The widely adopted choice for the normalization (indicated by $norm$ in the equation) is layer normalization \cite{ba2016layer} or RMSNorm \cite{zhang2019root}. Likewise, GELU \cite{hendrycks2016gaussian} for the activator function.

\section{Proposed Method}
Our Inductive Attention Model (IAM) combines an indexed memory with inductive attention lookup to summarize the video observations in a higher-order recurrent manner. %, as the function $\phi$ shown in eq~(\ref{eq:higher-order}).
An overview of the proposed architecture is shown in Figure~\ref{fig:recurrent_cell}.
\newline

\noindent \textbf{Indexed Memory.} 
Given the trajectory of previous recurrent state vectors, $h_{t-1}, h_{t-2}, \dots, h_{t-S}$, we introduce an indexed memory to cache the most recent state vectors. The indexed memory is a queue that stores past experiences as key-value pairs. For every timestep $t$, we project the anticipation prediction $\hat{y}_t \in \mathbb{R}^C$ ($C$ indicates the number of action classes) from a probability distribution to a low dimension vector of size $d/4$ by a function $E_K$ (a dense layer with ReLU), and assign the value by $h_t$.
\begin{equation}
    M_t = \{(E_K(\hat{y}_i), h_i)~for~i=t-1,\dots,t-S\} \label{eq:memory}
\end{equation}
where $S$ defines the maximal size of the queue (memory capacity). During the training, we stop the gradient of $\hat{y}_i$ in eq \eqref{eq:memory} to break the dependency from future timesteps.

The indexed memory follows the first-in-first-out (FIFO) update policy. We push the prediction $\hat{y}_t$ and recurrent state $h_t$ at every time $t$, and pop the oldest element in the queue that exceeds the memory capacity.
\newline

\noindent \textbf{Inductive Attention Model.}
For each timestep $t>0$, we have a previous prediction $\hat{y}_{t-1}$ and indexed memory $M_t = \{M_t^K, M_t^V\}$, with  $M_t^K, M_t^V$ presenting the keys and values stored in the memory.

We prepare $Q, K, V$ 
for inductive attention as follows. We project $\hat{y}_{t-1}$ from a vector of size $C$ down to a vector of size $d/4$ by a functions $E_Q$ (as on $\hat{y}_i$ with $E_K$)% for dense representation
. $M_t^K, M_t^V$ are assigned to $K$ and $V$, respectively. Inductive Attention, $IA$, is then defined as follows (for simplicity, we ignore the input embeddings in Multi-Head Attention, $MHA$):
\begin{align}
    IA(\hat{y}_{t-1},M_{t}) &\coloneqq FFN(MHA(Q_t, K_t, V_t)), \label{eq:ia}\\
    \nonumber\\
    where~Q_t &= E_Q(\hat{y}_{t-1}), \label{eq:q} \\
    K_t &= M^K_{t-1}, \label{eq:k} \\
    V_t &= M^V_{t-1}, \label{eq:v}
\end{align}
% The RMSNorm \cite{zhang2019root} is adopted as the normalization function in pre-norm style \cite{xiong2020layer} for attention usage (THIS IS NOT IN THE EQUATION. 

\begin{figure*}[t]
    \centering
    \includegraphics[width=1\textwidth]{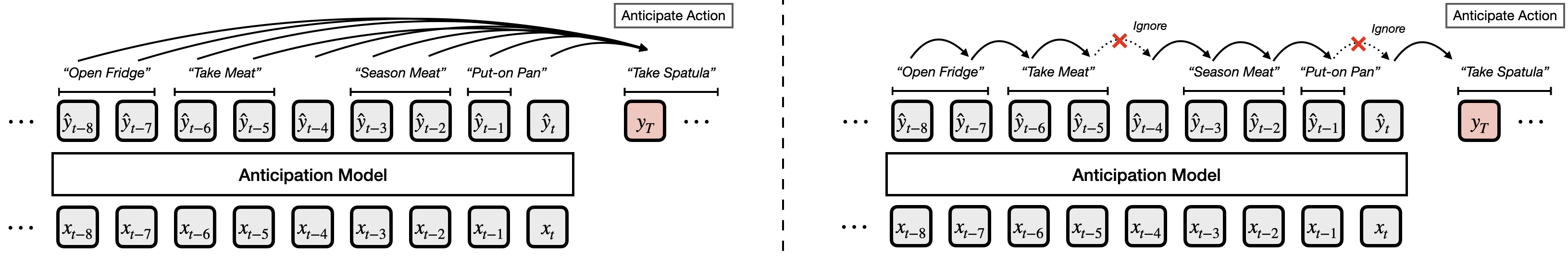}
    \caption{Supervision schemes with example activity. \textbf{Conventional learning (left)} uses the target action at the fixed time point as supervision. \textbf{Our learning scheme (right)} uses the next action, executed $\tau_a$ steps ahead in time, as supervision signal that may change at any step in time. This training scheme mimics the continuous inference scenario.}
    \label{fig:supervision_scheme}
\end{figure*}

\begin{figure}[t]
    \centering
    \includegraphics[width=1\columnwidth]{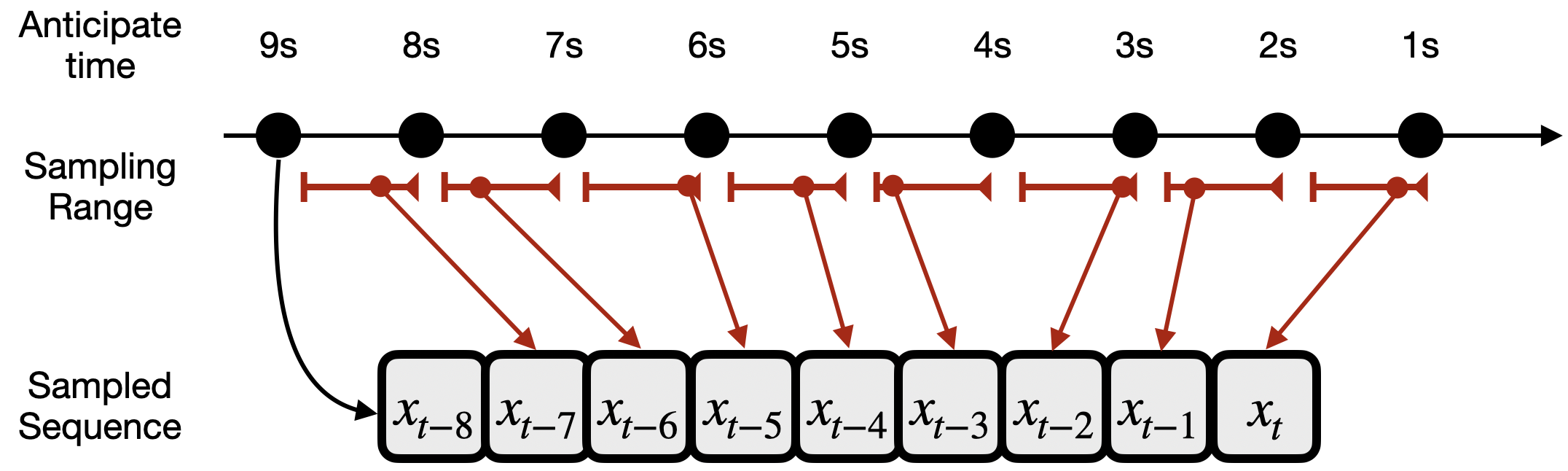}
    \caption{Jitter sampling augmentation replaces each video frame $x_t$ in the train sequence by randomly drawing a source frame from the video segment enclosed by $x_{t-1}$ and $x_t$. The first frame (i.e., $x_{t-8}$ in the figure) keeps intact to prevent crossing the video boundary.}
    \label{fig:augmentation}
\end{figure}

\begin{algorithm}[t]
\caption{Inductive Attention Model}\label{alg:iam}

\begin{algorithmic}
\State \textbf{Given} $E_x$ to project input to hidden state of size $d$; $E_K$ to compress vector of size $C$ to size $\frac{d}{4}$; $sg(.)$ is stop-gradient.
\State \textbf{Initialize} $M_0 \gets \varnothing$
\For{every receiving input $x_t$} \Comment{{\small\color{gray}$t$ for time index.}}
\State $e_t \gets E_x(x_t)$
\If{$t=0$}
\State $o_t \gets 0$
\Else
\State $o_t \gets IA(\hat{y}_{t-1}, M_{t})$ \Comment{{\small\color{gray}eq~(\ref{eq:ia}).}}
\EndIf
\State $g_t \gets \sigma(w_2^T(max(0, [o_t; e_t]w_1)))$
\State $h_t \gets g_t \cdot o_t + (1-g_t) \cdot e_t$
\State $\hat{y}_t \gets f_\theta(h_t)$
\State $M_{t+1}$ $\gets$ push $(E_K(sg(\hat{y}_t)), h_t)$ to $M_t$
\If{$M_{t+1}$ exceeds the maximal capacity}
\State pop the oldest element in $M_{t+1}$
\EndIf
\State \textbf{yield} $\hat{y}_t$
\EndFor
\end{algorithmic}
\end{algorithm}

Next, we compute the hidden state $h_t$ from attention and the frame input $x_t$ ($x_t$ is obtained from the video frame using a feature extractor). We also encode $x_t$ to a size $d$ vector by $E_x$, a dense layer with ReLU.
\begin{align}
    e_t &= E_x(x_t) \label{eq:encode_x}\\
    o_t &= IA(\hat{y}_{t-1}, M_{t}) \\ 
    g_t &= \sigma(w_2^T(max(0, [o_t; e_t]w_1))) \label{eq:gate} \\
    h_t &= g_t \cdot o_t + (1-g_t) \cdot e_t \label{eq:hidden_state}\\
    \hat{y}_t &= f_\theta(h_t) \label{eq:classify}
\end{align}
Here $f_\theta$ is an anticipation predictor, parameterized by weights $\theta$, which maps the hidden state into the action distribution. $g_t$ controls the mixture ratio between past experience (the output of inductive attention) and current frame input by feeding the concatenated $[o_t;e_t]$ to an MLP. The MLP uses $w_1,w_2$ to down-sample and up-sample the input by a factor of 4. The sigmoid function $\sigma(.)$ is to squash values into the range [0, 1]. We set $o_t=0$ for the initial step $t=0$, when memory is empty. Algorithm \ref{alg:iam} summarizes in pseudo-code the forwarding process.

Our model carries out a many-to-many association while computing the dot-product $Q^TK$ in eq~(\ref{eq:ia}). By using $Q\equiv\hat{y}_{t-1}$ and $K\equiv[\hat{y}_{t-2},\dots,\hat{y}_{t-S}]$, it estimates the pairwise correlation between previous predictions of all classes in $\hat{y}_*$.
\newline

\noindent \textbf{Training Objective.}
Prior works in action anticipation mostly adopted the training scheme from the action recognition setting (Figure~\ref{fig:supervision_scheme} left), where a model maps every timestep of an input segment to a global anticipation target. Although training this way might help capture and map long-range activity trajectories at once to the anticipation target, models could struggle to associate an early action of the input sequence to the future target if they are separated over a long period, e.g., tens of seconds. For instance, consider an activity containing the following actions: $"Open~Fridge" \rightarrow "Take~Meat" \rightarrow "Season~Meat \rightarrow "Put\text{-}on~Pan" \rightarrow "Take~Spatula"$. The target action $"Take~Spatula"$ will have a decreasing connection to its preceding actions that rarely exist between $"Open~Fridge"$ and $"Take~Spatula"$.

In our training scheme (Figure~\ref{fig:supervision_scheme} right), we treat every timestep equally and force the model always to predict the target action that happens in $\tau_a$ seconds. $\tau_a$ is the anticipation time: it indicates how many seconds after the last observation the action to anticipate will occur, and is defined by the benchmark (e.g., $\tau_a=1s$ in EPIC-KITCHENS-100). Indeed, the learned knowledge of $"Open~Fridge" \rightarrow "Take~Meat"$ (subsequent actions at close distance) is assumed to reflect stronger generalization than $"Open~Fridge" \rightarrow "Take~Spatula"$ (temporally far).
We furthermore ignore steps where the interval of length $\tau_a$ does not contain any action when computing the training loss.  Note that even though in our scheme we supervise the model not at segment level but at the frame level, the temporal context used to generate the prediction will still enable the model to accumulate enough evidence over time.
\newline

\noindent \textbf{Jitter Augmentation.}
\label{sec:jit_aug}
Rather than using the original frame rate in video modeling, a down-sampling is usually performed to lower the frame rate for training efficiency. According to this, we propose jitter augmentation that replaces each video frame $x_t$ in the train sequence by randomly drawing a source frame from the video segment enclosed by $x_{t-1}$ and $x_t$. For example, given a training video sampled at $F_{train}$ frame-per-seconds (fps) and a source device providing $F_{device}$ fps, we would have $F_{device} - F_{train}$ frames not being utilized at training time. Therefore, we can re-sample the frame index of a training sequence by offsetting $[-(F_{device} - F_{train}), 0]$ for augmentation. Figure~\ref{fig:augmentation} demonstrates the jitter augmentation.

\label{sec:exp}
\begin{table*}[t]
\caption{EPIC-Kitchens-100 validation result. The performance is measured in mean top-5 report (MT5R) at the $\tau_a=1s$. Individual actions (A), verbs (V), and nouns (N) accuracy is further presented. Rows which are highlighted in gray color indicate the methods are based on the same backbone features.}
    \vskip 5pt
    \centering
    \small
    \scalebox{0.79}{
    \begin{tabular}{l|l|l|c|ccc|ccc|ccc}
        \multirow{2}{*}{Methods} & \multirow{2}{*}{Modality} & \multirow{2}{*}{External Data} & Param & \multicolumn{3}{c}{Overall Classes} & \multicolumn{3}{c}{Unseen Classes} & \multicolumn{3}{c}{Tail Classes} \\
        \cline{5-13}
        & & & Counts & A & V & N & A & V & N & A & V & N \\
        \hline
        TempAgg \cite{sener2020temporal} & RGB, Obj, Flow, ROI & IN1K + EPIC bbox & - & 14.7 & 23.2 & 31.4 & 14.5 & 28.0 & 26.2 & 11.8 & 14.5 & 22.5 \\
        RULSTM \cite{furnari2020rulstm} & RGB, Obj, Flow & IN1K + EPIC bbox & - & 14.0 & 27.8 & 30.8 & 14.2 & 28.8 & 27.2 & 11.1 & 19.8 & 22.0 \\
        TSN-AVT+ \cite{girdhar2021anticipative} & RGB, Obj & IN21K + EPIC bbox & - & 14.8 & 25.5 & 31.8 & 11.5 & 25.5 & 23.6 & 12.6 & 18.5 & 25.8 \\
        AVT+ \cite{girdhar2021anticipative} & RGB, Obj & IN21K + EPIC bbox & - & 15.9 & 28.2 & 32.0 & 11.9 & 29.5 & 23.9 & 14.1 & 21.1 & 25.8 \\
        \hline
        chance & - & - & - & 0.2 & 6.4 & 2.0 & 0.5 & 14.4 & 2.9 & 0.1 & 1.6 & 0.2 \\
        \rowcolor{Gray}
        TempAgg \cite{sener2020temporal} & RGB & IN1K & - & 13.0 & 24.2 & 29.8 & 12.2 & 27.0 & 23.0 & 10.4 & 16.2 & 22.9 \\
        \rowcolor{Gray}
        RULSTM \cite{damen2022rescaling} & RGB & IN1K & - & 13.3 & 27.5 & 29.0 & - & - & - & - & - & - \\
        \rowcolor{Gray}
        HORST \cite{tai2021higher} & RGB & IN1K & - & 13.2 & 24.5 & 30.0 & - & - & - & - & - & - \\
        \rowcolor{Gray}
        MPNNEL-TB \cite{tai2022unified} & RGB & IN1K & - & 14.8 & 28.7 & 31.4 & - & - & - & - & - & - \\
        \rowcolor{Gray}
        TSN-AVT \cite{girdhar2021anticipative} & RGB & IN1K & - & 13.6 & 27.2 & 30.7 & - & - & - & - & - & - \\ 
        irCSN152-AVT \cite{girdhar2021anticipative} & RGB & IG65M & - & 12.8 & 25.5 & 28.1 & - & - & - & - & - & - \\
        AVT \cite{girdhar2021anticipative} & RGB & IN1K & 378M & 13.4 & 28.2 & 29.3 & - & - & - & - & - & - \\
        AVT \cite{girdhar2021anticipative} & RGB & IN21+1K & 378M & 14.4 & 28.7 & 32.3 & - & - & - & - & - & - \\
        AVT \cite{girdhar2021anticipative} & RGB & IN21K & 378M & 14.9 & 30.2 & 31.7 & - & - & - & - & - & - \\
        \rowcolor{Gray}
        TSN-DCR-LSTM \cite{xu2022learning} & RGB & IN1K & 14M & 14.5 & 27.9 & 28.0 & - & - & - & - & - & - \\
        TSM-DCR-LSTM \cite{xu2022learning} & RGB & IN1K & 20M & 15.2 & 28.4 & 28.5 & - & - & - & - & - & - \\
        \rowcolor{Gray}
        TSN-DCR \cite{xu2022learning} & RGB & IN1K & 78M & 14.6 & 31.0 & 31.1 & - & - & - & - & - & - \\
        TSM-DCR \cite{xu2022learning} & RGB & IN1K & 84M & 16.1 & 32.6 & 32.7 & - & - & - & - & - & - \\
        MeMViT 16x4 \cite{wu2022memvit} & RGB & K400 & 59M & 15.1 & \textbf{32.8} & 33.2 & 9.8 & 27.5 & 21.7 & 13.2 & 26.3 & 27.4 \\
        MeMViT 32x3 \cite{wu2022memvit} & RGB & K700 & 212M & 17.7 & 32.2 & 37.0 & 15.2 & 28.6 & 27.4 & 15.5 & 25.3 & 31.0 \\
        \hline
        \rowcolor{Gray}
        \textbf{TSN-IAM (Ours)} & RGB & IN1K & 42M & 17.5 & 32.2 & 35.7 & 11.9 & 31.4 & 24.9 & 16.8 & \textbf{26.9} & 31.8 \\
        \textbf{ConvNeXt-IAM (Ours)} & RGB & IN22K & 142M & 17.6 & 31.4 & 36.2 & 12.0 & 33.0 & 25.0 & 17.1 & 26.0 & 32.0 \\
        \textbf{Swin-IAM (Ours)} & RGB & IN22K & 141M & \textbf{18.1} & 32.1 & \textbf{37.2} & \textbf{16.6} & \textbf{34.6} & \textbf{27.9} & \textbf{17.6} & 26.7 & \textbf{33.0}
    \end{tabular}
    }
    \label{tab:ek100}
\end{table*}

%                                           0
% Overall Mean Top-5 Recall Verb    32.116061
%                           Noun    37.241800
%                           Action  18.123050
% Unseen Mean Top-5 Recall  Verb    34.558749
%                           Noun    27.943476
%                           Action  16.591495
% Tail Mean Top-5 Recall    Verb    26.748763
%                           Noun    32.997527
%                           Action  17.578885

%                                           0
% Overall Mean Top-5 Recall Verb    32.161462
%                           Noun    35.716873
%                           Action  17.515462
% Unseen Mean Top-5 Recall  Verb    31.405821
%                           Noun    24.879228
%                           Action  11.856667
% Tail Mean Top-5 Recall    Verb    26.963574
%                           Noun    31.814839
%                           Action  16.848655

%                                           0
% Overall Mean Top-5 Recall Verb    31.355028
%                           Noun    36.203101
%                           Action  17.589324
% Unseen Mean Top-5 Recall  Verb    32.952965
%                           Noun    25.030642
%                           Action  12.041013
% Tail Mean Top-5 Recall    Verb    25.997828
%                           Noun    31.966836
%                           Action  17.145702
\begin{table}[t]
\caption{EPIC-Kitchens-55 validation results at the $\tau_a=1s$. The top-1/top-5 action accuracy and top-5 action recall are summarized. The highlighted rows indicate the fair comparison under the same backbone usage. All methods are done in RGB modality.}
    \vskip 5pt
    \centering
    \small
    \scalebox{0.78}{
    \begin{tabular}{l|l|l|cc|c}
        \multirow{2}{*}{Methods} & \multirow{2}{*}{Backbone} & External & \multirow{2}{*}{Top-1} & \multirow{2}{*}{Top-5} & \multirow{2}{*}{Recall} \\
         & & Data & & & \\
        \hline
        \rowcolor{Gray}
        RULSTM \cite{furnari2020rolling} & TSN & IN1K & 13.1 & 30.8 & 12.5 \\
        \rowcolor{Gray}
        TempAgg \cite{sener2020temporal} & TSN & IN1K & 12.3 & 28.5 & 13.1 \\
        \rowcolor{Gray}
        ImagineRNN \cite{wu2020learning} & TSN & IN1K & 13.7 & 31.6 & - \\
        \rowcolor{Gray}
        SRL \cite{qi2021self} & TSN & IN1K & - & 31.7 & 13.2 \\
        \rowcolor{Gray}
        HORST \cite{tai2021higher} & TSN & IN1K & 12.8 & 31.6 & 12.2 \\
        \rowcolor{Gray}
        MPNNEL-TB \cite{tai2022unified} & TSN & IN1K & 13.8 & 32.0 & 13.6 \\
        \rowcolor{Gray}
        AVT-h \cite{girdhar2021anticipative} & TSN & IN1K & 13.1 & 28.1 & 13.5 \\
        AVT-h \cite{girdhar2021anticipative} & AVT-b & IN21+1K & 12.5 & 30.1 & 13.6 \\
        AVT-h \cite{girdhar2021anticipative} & irCSN152 & IG65M & 14.4 & 31.7 & 13.2 \\
        \rowcolor{Gray}
        DCR \cite{xu2022learning} & TSN & IN1K & 13.6 & 30.8 & - \\
        DCR \cite{xu2022learning} & irCSN152 & IG65M & 15.1 & \textbf{34.0} & - \\
        DCR \cite{xu2022learning} & TSM & IN1K & \textbf{16.1} & 33.1 & - \\
        \hline
        \rowcolor{Gray}
        \textbf{IAM (Ours)} & TSN & IN1K & 13.5 & 32.1 & 14.3 \\
        \textbf{IAM (Ours)} & ConvNeXt & IN22K & 13.3 & 32.6 & 15.1 \\
        \textbf{IAM (Ours)} & Swin & IN22K & 14.2 & \textbf{34.0} & \textbf{16.1}
    \end{tabular}
    }
    \label{tab:ek55}
\end{table}

%                                   0
% Verb   Top-1 Accuracy     32.300262
%        Top-5 Accuracy     79.110485
%        Mean Top-5 Recall  41.406985
% Noun   Top-1 Accuracy     21.392634
%        Top-5 Accuracy     48.480580
%        Mean Top-5 Recall  48.904242
% Action Top-1 Accuracy     13.463474
%        Top-5 Accuracy     32.119139
%        Mean Top-5 Recall  14.291740

% swin
%                                   0
% Verb   Top-1 Accuracy     33.226001
%        Top-5 Accuracy     80.338096
%        Mean Top-5 Recall  44.031761
% Noun   Top-1 Accuracy     23.244114
%        Top-5 Accuracy     52.062789
%        Mean Top-5 Recall  54.275817
% Action Top-1 Accuracy     14.187965
%        Top-5 Accuracy     34.030992
%        Mean Top-5 Recall  16.120241

%  convnext
%                                   0
% Verb   Top-1 Accuracy     33.407124
%        Top-5 Accuracy     79.633729
%        Mean Top-5 Recall  40.497747
% Noun   Top-1 Accuracy     22.137251
%        Top-5 Accuracy     51.137050
%        Mean Top-5 Recall  51.534500
% Action Top-1 Accuracy     13.282351
%        Top-5 Accuracy     32.602133
%        Mean Top-5 Recall  15.115899
\begin{table}[t]
\caption{EGTEA Gaze+ validation results on the split 1 at the $\tau_a=0.5s$. We reported the top-1 accuracy and mean top-1 recall of each individual action (A), verb (V) and noun (N). The methods highlighted in background are under the same backbone usage. All methods are based on RGB modality.}
    \small
    \centering
    \scalebox{0.8}{
    \begin{tabular}{l|ccc|ccc}
        \multirow{2}{*}{Methods} & \multicolumn{3}{c}{Top-1 Acc} &  \multicolumn{3}{c}{Mean Top-1 Recall} \\
        \cline{2-7}
        & A & V & N & A & V & N \\ 
        \hline
        I3D-Res50 \cite{carreira2017quo} & 34.8 & 48.0 & 42.1 & 23.2 & 31.3 & 30.0 \\
        FHOI \cite{liu2020forecasting} & 36.6 & 49.0 & 45.5 & 32.5 & 32.7 & 25.3 \\
        \rowcolor{Gray}
        TSN-AVT-h \cite{girdhar2021anticipative} & 39.8 & 51.7 & 50.3 & 28.3 & 41.2 & 41.4 \\
        AVT \cite{girdhar2021anticipative} & 43.0 & 54.9 & 52.2 & 35.2 & \textbf{49.9} & 48.3 \\
        \hline
        \rowcolor{Gray}
        \textbf{TSN-IAM (Ours)} & 43.5 & 54.3 & 52.2 & 35.5 & 43.8 & 46.6 \\
        \textbf{ConvNeXt-IAM (Ours)} & 44.6 & 54.5 & 53.1 & 36.3 & 42.6 & 45.3 \\
        \textbf{Swin-IAM (Ours)} & \textbf{45.4} & \textbf{55.9} & \textbf{54.3} & \textbf{37.4} & 46.5 & \textbf{49.3}
    \end{tabular}
    }
    \label{tab:gaze}
\end{table}

% egtea_norefresh                      
% Verb   Top-1 Accuracy     54.302671
%        Top-5 Accuracy     93.818002
%        Mean Top-1 Recall  43.752578
% Noun   Top-1 Accuracy     52.176063
%        Top-5 Accuracy     81.998022
%        Mean Top-1 Recall  46.631563
% Action Top-1 Accuracy     43.471810
%        Top-5 Accuracy     72.007913
%        Mean Top-1 Recall  35.547108

% swin backbone
% Verb   Top-1 Accuracy     55.934718
%        Top-5 Accuracy     93.125618
%        Mean Top-1 Recall  46.569584
% Noun   Top-1 Accuracy     54.302671
%        Top-5 Accuracy     82.096934
%        Mean Top-1 Recall  49.330930
% Action Top-1 Accuracy     45.351137
%        Top-5 Accuracy     73.392681
%        Mean Top-1 Recall  37.443700

% convnext backbone
% Verb   Top-1 Accuracy     54.549951
%        Top-5 Accuracy     94.362018
%        Mean Top-1 Recall  42.557471
% Noun   Top-1 Accuracy     53.115727
%        Top-5 Accuracy     84.569733
%        Mean Top-1 Recall  45.294516
% Action Top-1 Accuracy     44.609298
%        Top-5 Accuracy     74.975272
%        Mean Top-1 Recall  36.253030

\section{Experiments}
We benchmark our proposed method in three popular egocentric video datasets for action anticipation. The ablation study is presented to validate the design choice in our proposed model.

\subsection{Implementation Details}
In all experiments, we adopted TSN \cite{furnari2020rulstm}, ConvNeXt \cite{liu2022convnet}, and Swin \cite{liu2021swin} as backbone variants in our model. In all evaluation tables we highlight in gray the methods based on the same feature set for fair comparisons. 

We trained our model using AdamW optimizer \cite{loshchilov2017decoupled} with a learning rate set to 2e-4 and a cosine decay scheduler. We set weight decay to 1e-2 and exclude biases and normalization layers from weight decay. We use batch size 128 and train for 50 epochs on a single NVIDIA RTX 3090 GPU with automatic mixed precision enabled. Throughout this study, our model contains a single IAM layer instantiated with hidden size set to $d=2048$ for all the internal variables defined in eq~(\ref{eq:encode_x}-\ref{eq:hidden_state}). We used dropout rate 0.6. Other hyperparameters are listed individually for each dataset.

\subsection{Datasets}
\noindent \textbf{EPIC-Kitchens-100 (EK100) \cite{damen2022rescaling}:} 
EK100 is a large-scale egocentric video dataset containing 100 hours of recordings. It includes 3806 action labels consisting of 97 verbs and 300 nouns, which splits into 67217 segments for training and 9668 for validation. Similar to \cite{wu2022memvit} that used an equalized loss for addressing the class imbalance, we rescale the loss weightings proportional to the inverse counts of class occurrences in the training set. We also apply the jitter augmentation described in section \ref{sec:jit_aug}. Mean top-5 recall (MT5R) for action, verb, and noun is measured at anticipation interval $\tau_a=1s$. We sampled each clip of the dataset at 1 fps to obtain 30 frames.
\newline

\noindent \textbf{EPIC-Kitchens-55 (EK55) \cite{damen2018scaling}:} 
EK55 contains 55 hours of egocentric video recordings with 2513 action classes comprising 125 verbs and 352 nouns. It includes 23492 segments for training and 4979 for validation. We did not apply the jitter augmentation. Instead, the label smoothing \cite{muller2019does} of ratio 0.6 is adopted. Top-1/5 action accuracy and mean top-5 action recall (MT5R) at $\tau_a=1s$ are reported. We sampled the dataset at 1 fps to obtain 10 frames.
\newline

\noindent \textbf{EGTEA Gaze+ \cite{li2018eye}:}
EGTEA Gaze+ is another egocentric video dataset containing more than 28 hours of video recordings. The dataset comes with 10321 segments annotated with 106 unique action classes. The video segments are further split into 3 folds. We followed the experiment setup in \cite{liu2020forecasting, girdhar2021anticipative}, which evaluates on split 1, with 8299 clips for training and 2022 for validation. Anticipation accuracy is assessed at $\tau_a=0.5s$ for top-1 action accuracy and mean top-1 action recall. We adopted label smoothing with a ratio of 0.4 and sampled the dataset at 2 fps to obtain 10 frames.

\subsection{EK100 Action Anticipation Result}
We compared our proposed model (IAM) with prior works on the EK100 dataset and summarized the results in Table~\ref{tab:ek100}. In addition, methods using pre-extracted TSN features from the dataset developers are highlighted (in gray background color) for fair comparisons. Our proposed model outperforms those methods by a large margin. We obtained +2.7 gains in mean top-5 recall in overall action accuracy, -0.3 in unseen action classes, and +6.4 in tail action classes. Although slightly lower unseen action accuracy has been observed, much better verb (+4.4) and noun (+1.9) predictions were achieved. In brief, using pre-extracted TSN features significantly improves over some strong baselines, including the previous top-rank competition winner AVT, and is competitive with the state-of-the-art MeMViT.

Furthermore, by replacing the TSN with recent ConvNeXt (convolutional-based) and Swin (transformer-based) backbones, IAM outperforms MeMViT by a significant margin: +0.4 in overall classes, +1.4 in unseen classes, and +2.1 in tail classes. Moreover, our method is with fewer model parameters. Figure~\ref{fig:param_plot} shows the efficiency and efficacy of the proposed IAM on the EK100 evaluation.

\subsection{EK55 Action Anticipation Result}
Table~\ref{tab:ek55} summarizes the performance comparisons of our model and prior works on the EK55 benchmark. TempAgg \cite{sener2020temporal} leveraged multi-scale temporal information and aggregate for the action anticipation; SRL \cite{qi2021self} enhanced the recurrent update by re-attend the past observations; HORST \cite{tai2021higher}, MPNNEL-TB \cite{tai2022unified} consider the spatial-temporal information in recurrent modeling; ImagineRNN \cite{wu2020learning} generate the future frame to guide the prediction; DCR \cite{xu2022learning} introduces a curriculum learning scheme and additional pretraining improvements.

The evaluation results show our model outperforms others on the top-5 action accuracy and mean top-5 recall based on the same TSN baseline. We achieve further performance gains while using ConvNeXt and Swin as feature extractors. Like the EK100 evaluation, our model with the Swin backbone achieved the best overall performance. Note that AVT-h on the ViT-based AVT-b backbone does not perform better in accuracy, unlike in EK100, than AVT-h with the convolutional backbone irCSN152 pre-trained on IG65M, indicating the AVT does not produce consistent scores by one specific backbone across different datasets. On the other hand, DCR achieved higher top-1 accuracy, a possible gain via their curriculum learning from the future frames. This observation exposes the emergence of an appropriate pretraining method for video predictive tasks. We mark it as a future direction.

\newcommand{\PreserveBackslash}[1]{\let\temp=\\#1\let\\=\temp}
\newcolumntype{C}[1]{>{\PreserveBackslash\centering}p{#1}}
\newcolumntype{R}[1]{>{\PreserveBackslash\raggedleft}p{#1}}
\newcolumntype{L}[1]{>{\PreserveBackslash\raggedright}p{#1}}

% \begin{table}[t]
% \caption{Ablation on using different input lengths. The experiment is done on EPIC-Kitchens-100 dataset and measured in mean top-5 recall (MT5R). {\color{red}{could only contain TSN results to align with other ablation. (YES ALL ABLATIONS TSN)}}}
%     \vskip 5pt
%     \centering
%     \small
%     \scalebox{0.95}{
%     \begin{tabular}{C{1.5cm}|C{1.7cm}|C{1.7cm}|C{1.7cm}}
%         Input & \multicolumn{3}{c}{MT5R (\%)} \\
%         \cline{2-4}
%         Length & TSN & ConvNeXt & Swin \\
%         \hline
%         10s & 16.9 & 16.9 & 17.2 \\
%         20s & 17.5 & 17.4 & 18.0 \\
%         30s & 17.5 & 17.6 & 18.1 \\
%         40s & 17.3 & 16.9 & 18.0 \\
%         50s & 16.7 & 17.5 & 17.9 \\
%         60s & 16.7 & & 18.0 \\
%     \end{tabular}
%     }
%     \label{tab:ablation_context}
%     % \vspace{-.2in}
% \end{table}

\begin{table}[t]
\caption{Ablation on using different input lengths. The experiment is done on EPIC-Kitchens-100 dataset and measured in mean top-5 recall (MT5R).}
    \vskip 5pt
    \centering
    \small
    \scalebox{0.95}{
    \begin{tabular}{l|c|c|c|c|c|c|c}
        Input Length & 10s & 20s & 30s & 40s & 50s & 60s \\
        \hline
        MT5R (\%) & 16.9 & \textbf{17.5} & \textbf{17.5} & 17.3 & 16.7 & 16.7 \\
    \end{tabular}
    }
    \label{tab:ablation_context}
    % \vspace{-.2in}
\end{table}

\begin{table}[t]
\caption{Ablation on jitter sampling augmentation. The accuracy and recall defined in each dataset are different, in which action top-5 is used for EK55/100, and top-1 for EGTEA Gaze+.}
    \vskip 5pt
    \centering
    \small
    \scalebox{0.8}{
    \begin{tabular}{l|l|c|c|c}
        \multirow{2}{*}{Dataset} & \multirow{2}{*}{Input} & Jitter & Action & Action \\
         & & Sampling & Accuracy & Recall \\
        \hline
        \multirow{2}{*}{EK100} & \multirow{2}{*}{30 frames @ 1fps} & \checkmark & 24.7 & \textbf{17.5} \\
         &  &  & \textbf{27.3} & 16.9 \\
        \hline
        % \hline
        \multirow{2}{*}{EK55} & \multirow{2}{*}{10 frames @ 1fps} & \checkmark & 31.6 & 13.9 \\
         &  &  & \textbf{32.1} & \textbf{14.3} \\
        \hline
        % \hline
        \multirow{2}{*}{EGTEA Gaze+} & \multirow{2}{*}{10 frames @ 2fps} & \checkmark & 43.2 & \textbf{36.0} \\
         &  &  & \textbf{43.5} & 35.5 \\
    \end{tabular}
    }
    \label{tab:ablation_aug}
    % \vspace{-.2in}
\end{table}

\subsection{EGTEA Gaze+ Action Anticipation Result}
We also evaluated our model on the EGTEA Gaze+ egocentric video dataset. Table~\ref{tab:gaze} includes the task baselines, where I3D-Res50 \cite{carreira2017quo} proposed a clip-based 3D convolutions model; FHOI \cite{liu2020forecasting} utilizes the hand movements using a 2D resnet-50 backbone; and the previous state-of-the-art AVT. The results show that models based on sequential or recurrent networks outperform clip-based models, confirming the conclusions drawn in \cite{wang2018eidetic, su2020convolutional}. 

Based on the same pre-extracted TSN features, our model significantly improves top-1 action accuracy and top-1 action recall over the TSN-AVT. Furthermore, we achieve state-of-the-art action accuracy when using the recent ConvNeXt or Swin transformer as backbone.

\subsection{Ablation Study}
This section analyzes the optimal input length for IAM. In addition, the effect of jitter augmentation and the selection of the query are also covered. All the experiments are done on EK100 using the pre-extracted baseline TSN features.
\newline

\noindent \textbf{Input Length.}
We first evaluated the performance using different lengths of inputs, ranging from 10 to 60 seconds. The longer input length implies more pre-action trajectories in each training sequence. Table~\ref{tab:ablation_context} shows that our model reaches peak performance when fed with 20 to 30 seconds inputs and shows degradation when either increasing or decreasing the length. This result reveals that the model does not always have increasing performance while extending the contexts. A similar conclusion can also be found in the preliminary study \cite{furnari2020rolling}. However, the optimal input length for our model is significantly longer than earlier recurrent models have been reported. This evidence demonstrates that our higher-order inductive attention can effectively utilize historical experience in long video observations. In addition, compared with the previous higher-order recurrent networks, we significantly extended from 3-8 higher-order states to the 30 without suffering from gradient instability during learning. On the other hand, \cite{wu2022memvit} suggested taking long video inputs (longer than a minute) can improve the recognition performance of models. Our model based on recurrent mechanisms can reach equal or even better accuracy in the length of the same or shorter input with higher efficiency. This finding demonstrates the difference between sequential and recurrent modeling.

We analyze the memory footprint of our proposed indexed memory, which caches up to $S$ elements of hidden dimension size $d$. Each element needs $d/4$ for key and $d$ for value, totalling $(5/4)dS$. Consider our model with $S=30$ and $d=2048$. Under a single precision floating point, it only occupies about 300 KB during runtime inference, showing the resource impact of memory capacity is negligible in practice.
\newline

\noindent \textbf{Jitter Augmentation.}
We validate the jitter augmentation described in section \ref{sec:jit_aug} on different datasets in Table~\ref{tab:ablation_aug}. Overall, the proposed jitter augmentation increased the overall action recall (for 2 out of the 3 entries) but at the cost of accuracy (for 3 out of 3). The jitter augmentation varied the sampling rate during training, which is equivalent to randomly increasing  $\tau_a$ to $\widehat{\tau_a} \in [\tau_a, 2\tau_a)$. This makes the task more challenging, and the accuracy of the action is adversely affected. On the other hand, it encourages the model to consider more diverse responses. According to the trade-off between precision and recall, we suggest optionally applying jitter augmentation for the needs of different scenarios.
\newline

\begin{table}[t]
\caption{EK100 performance comparison using different choices for the query. The numbers are measured in mean top-5 recall (MT5R) for actions (A), verbs (V), nouns (N).}
    \vskip 5pt
    \centering
    \small
    \scalebox{0.7}{
    \begin{tabular}{l|ccc|ccc|ccc}
        \multirow{2}{*}{Query} & \multicolumn{3}{c}{Overall Classes} & \multicolumn{3}{c}{Unseen Classes} & \multicolumn{3}{c}{Tail Classes} \\
        \cline{2-10}
         & A & V & N & A & V & N & A & V & N \\
        \hline
        Frame Input & 17.0 & 31.1 & \textbf{35.7} & 11.1 & 27.4 & \textbf{24.9} & 16.4 & 25.6 & \textbf{32.1} \\
        Last Pred. & \textbf{17.5} & \textbf{32.2} & \textbf{35.7} & \textbf{11.9} & \textbf{31.4} & \textbf{24.9} & \textbf{16.8} & \textbf{26.9} & 31.8 \\
        \hline
        $\Delta$ & {\color{teal}+0.5} & {\color{teal}\textbf{+1.1}} & {\color{gray}+0.0} & {\color{teal}+0.8} & {\color{teal}\textbf{+4.0}} & {\color{gray}+0.0} & {\color{teal}+0.4} & {\color{teal}\textbf{+1.3}} & {\color{purple}-0.3}
    \end{tabular}
    }
    \label{tab:ek100_supp}
\end{table}

\begin{table}[t]
\caption{EK55 performance comparison using different query choices, in which frame features or previous predictions in our model.}
    \vskip 5pt
    \centering
    \small
    \scalebox{0.85}{
    \begin{tabular}{l|cc|c}
        Query & Top-1 & Top-5 & Recall \\
        \hline
        Frame Input & 13.2 & 31.4 & 13.9 \\
        Last Pred. & \textbf{13.5} & \textbf{32.1} & \textbf{14.3} \\
        \hline
        $\Delta$ & {\color{teal}+0.3} & {\color{teal}+0.7} & {\color{teal}+0.4}
    \end{tabular}
    }
    \label{tab:ek55_supp}
\end{table}

\begin{table}[t]
\caption{EGTEA Gaze+ performance comparison using different query choices, in which frame features or previous predictions in our model.}
    \vskip 5pt
    \small
    \centering
    \scalebox{0.85}{
    \begin{tabular}{l|ccc|ccc}
        \multirow{2}{*}{Query} & \multicolumn{3}{c}{Top-1 Acc} &  \multicolumn{3}{c}{Mean Top-1 Recall} \\
        \cline{2-7}
         & A & V & N & A & V & N \\ 
        \hline
        Frame Input & 43.0 & 54.2 & 51.7 & 34.5 & \textbf{44.9} & 45.8 \\
        Last Pred. & \textbf{43.5} & \textbf{54.3} & \textbf{52.2} & \textbf{35.5} & 43.8 & \textbf{46.6} \\
        \hline
        $\Delta$ & {\color{teal}+0.5} & {\color{teal}+0.1} & {\color{teal}+0.5} & {\color{teal}\textbf{+1.0}} & {\color{purple}-1.1} & {\color{teal}+0.8}
    \end{tabular}
    }
    \label{tab:gaze_supp}
\end{table}

\noindent \textbf{About Query.}
Finally, we show that using task prediction as the query in inductive attention can improve overall performance. 

Table~\ref {tab:ek100_supp} compares the results on the EK100 dataset. In the experiment with frame features as the query, we replace the prediction $\hat{y}_{t-1}$ in eq.~(\ref{eq:ia}) with frame input $x_t$ and remove the encoding modules ($E_Q, E_K$ in eq.~(\ref{eq:q})(\ref{eq:memory})). Using the last prediction as the query can achieve overall consistent improvement. A slight degradation on the tail class nouns is found, which may be attributed to the fact that frame input features contain more spatial information than task predictions. Overall, inductive attention with the last prediction as query improves over the verb classes by a margin.

Similarly, the general improvements using the last prediction query are shown in Tables~\ref{tab:ek55_supp} and \ref{tab:gaze_supp}. By combining our higher order recurrent model design with self-attention  and utilizing previous prediction as the query, IAM obtains state-of-the-art performance in three action anticipation benchmarks.

\section{Conclusion}
\label{sec:conclude}
This paper introduced IAM, an Inductive Attention Model for video action anticipation. IAM leverages a higher-order recurrent design to summarize the temporal information efficiently, and induces from historical moments stored in the indexed memory to infer the future action. Utilizing previous predictions as the query in inductive attention allows the proposed model to handle long videos and set new state-of-the-art accuracy on large-scale egocentric video datasets. Our method delivers higher accuracy and efficiency while requiring fewer model parameters than previous works.

%%%%%%%%% REFERENCES
{\small
\bibliographystyle{ieee_fullname}
\bibliography{egbib}
}

\clearpage
% \documentclass[10pt,twocolumn,letterpaper]{article}
% \usepackage{iccv}

% % Include other packages here, before hyperref.
% \usepackage{graphicx}
% \usepackage{amsmath}
% \usepackage{amssymb}
% \usepackage{booktabs}
% \usepackage{multirow}
% \usepackage{subfigure}

% \usepackage{color, colortbl}
% \definecolor{Gray}{gray}{0.9}
% \usepackage{xcolor}

% \usepackage[pagebackref=true,breaklinks=true,letterpaper=true,colorlinks,bookmarks=false]{hyperref}

% % Support for easy cross-referencing
% \usepackage[capitalize]{cleveref}
% \crefname{section}{Sec.}{Secs.}
% \Crefname{section}{Section}{Sections}
% \Crefname{table}{Table}{Tables}
% \crefname{table}{Tab.}{Tabs.}

% \usepackage{algorithm}
% \usepackage{algpseudocode}
% % \usepackage{algorithm2e}

% \usepackage{mathtools}

% \iccvfinalcopy % *** Uncomment this line for the final submission

% \def\iccvPaperID{7047} % *** Enter the ICCV Paper ID here
% \def\httilde{\mbox{\tt\raisebox{-.5ex}{\symbol{126}}}}

% % Pages are numbered in submission mode, and unnumbered in camera-ready
% \ificcvfinal\pagestyle{empty}\fi

% \begin{document}

% \title{Inductive Attention for Video Action Anticipation\\Supplementary Material}

% \maketitle

\appendix

\setcounter{figure}{4}
\setcounter{table}{8}

\section{Input Preprocessing}
For the TSN baseline used in three action anticipation benchmarks of Table 1,2,3 in the main paper, we downloaded the pre-extracted features from the official release of the EPIC-Kitchens websites\footnote{https://github.com/epic-kitchens/C3-Action-Anticipation}. TSN is based on the BN-Inception backbone and extracts RGB frame input in size (456, 256). Swin and Convnext implementation and pretrained weights used to present the results in Table 1,2,3 are inherited from the open-source implementation\footnote{https://github.com/huggingface/pytorch-image-models, v0.5.4}. The frame inputs for Swin and ConvNeXt are (224, 224) in size, with values rescaled to the range [-1, 1].

\section{Qualitative Analysis}
We draw success and failure cases of our model from EPIC-Kitchens-55/100 and EGTEA Gaze+ datasets for qualitative analysis. We illustrated four video clips, labeled \textbf{(a)} to \textbf{(d)}, for the case that the ground truth falls in the model top-5 predictions at the last frame. Also, another four examples, labeled \textbf{(e)} to \textbf{(h)}, failed to anticipate the target action. We presented the last eight frames for each video sample and the corresponding top-5 predictions per frame. The figures are best viewed horizontally and zoom-in.
\newline

\begin{figure*}
    \centering
    \includegraphics[width=0.89\textwidth]{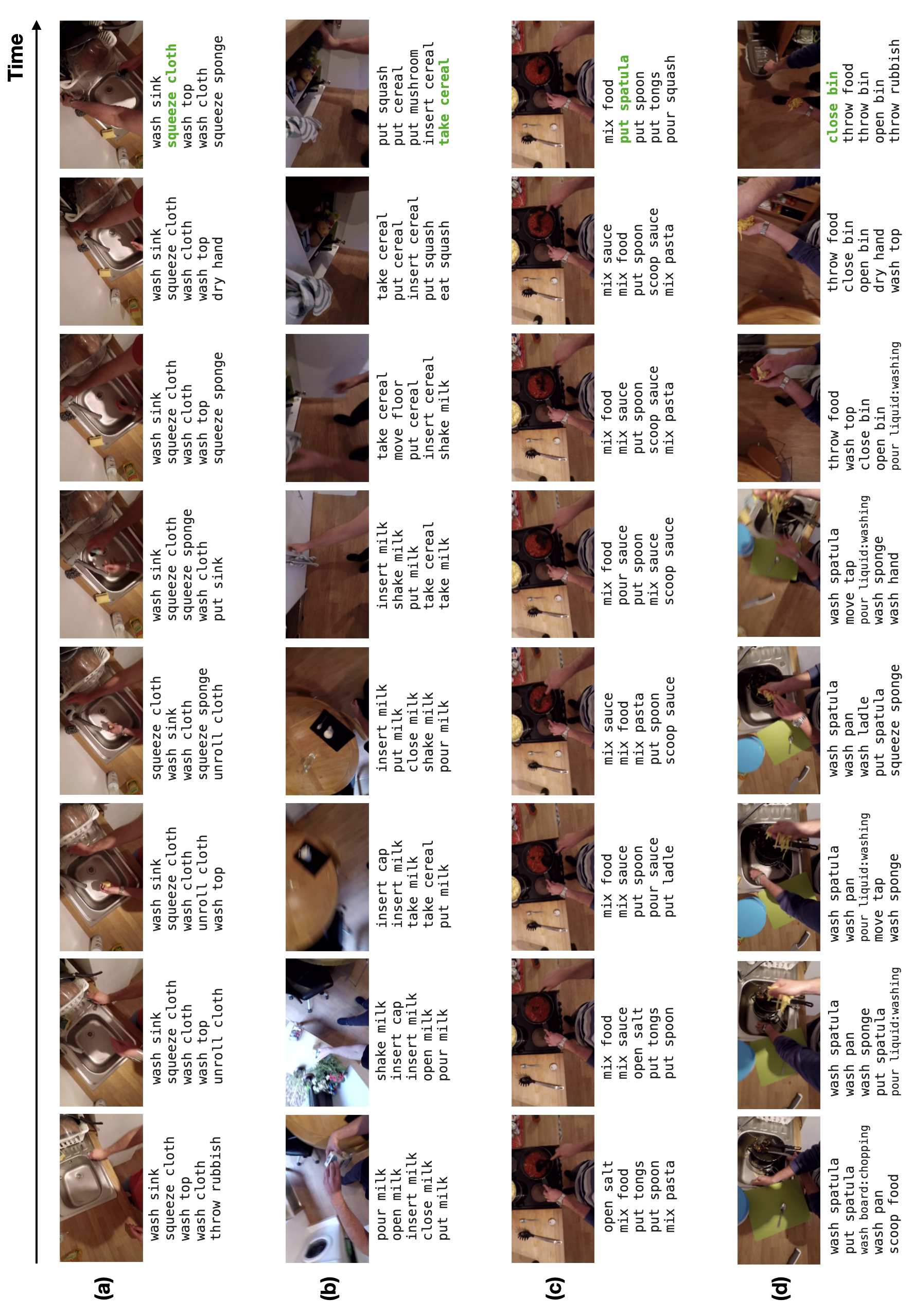}
    \caption{Some \textbf{correct} predictions illustrates on the \textbf{EPIC-Kitchens-100} validation set. The figure shows four video clips (the last eight frames presented) and the corresponding top-5 action anticipations ($\tau_a=1s$). The ground truth is highlighted in bold green.}
    \label{fig:eg100_correct}
\end{figure*}

\begin{figure*}
    \centering
    \includegraphics[width=0.89\textwidth]{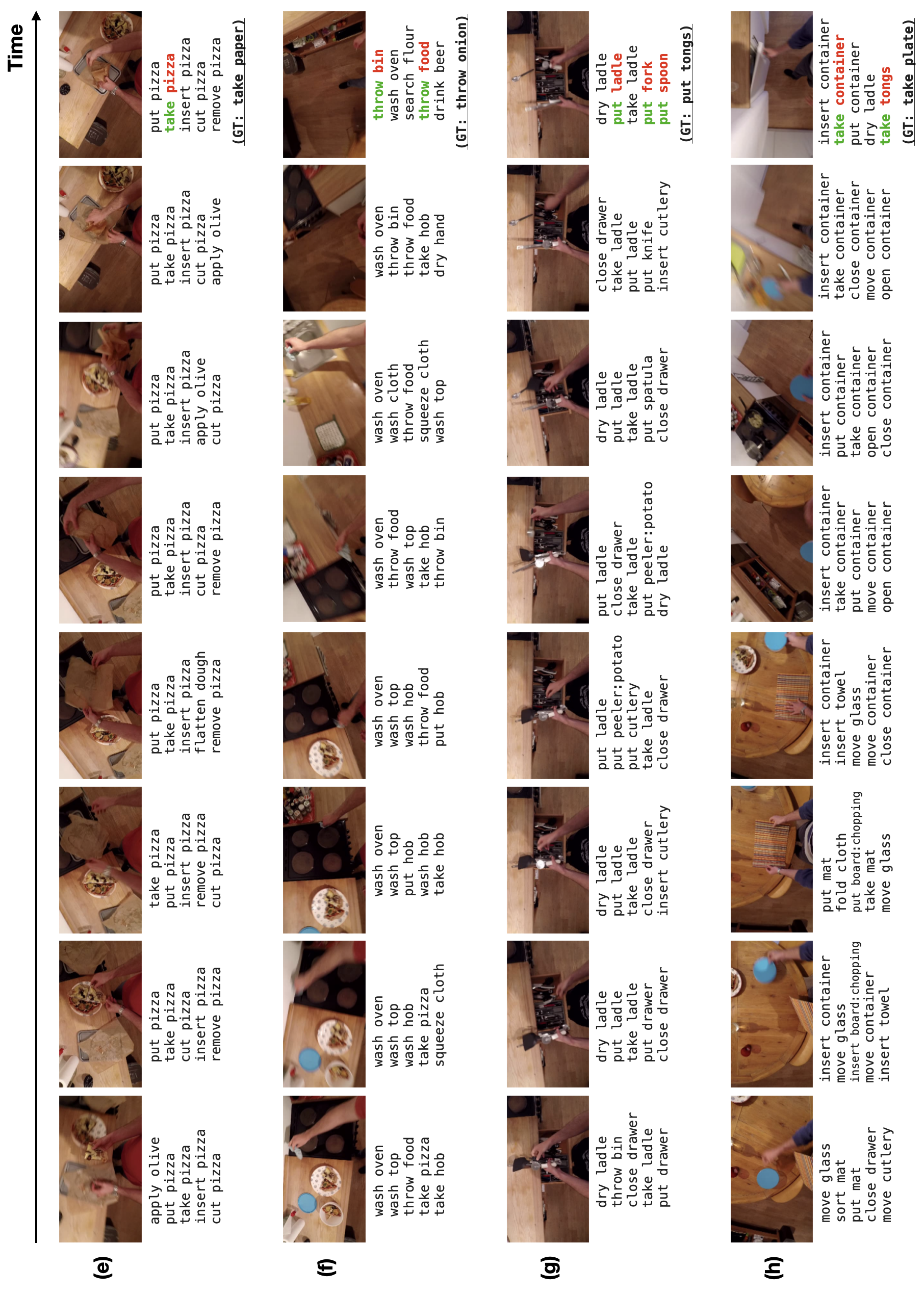}
    \caption{Some \textbf{incorrect} predictions illustrates on the \textbf{EPIC-Kitchens-100} validation set. The figure shows four video clips (the last eight frames presented) and the corresponding top-5 action anticipations ($\tau_a=1s$). The ground truth is revealed at the end. The correct verb and noun of action are highlighted in green.}
    \label{fig:eg100_incorrect}
\end{figure*}

\noindent \textbf{EPIC-Kitchens-100.}
Figure \ref{fig:eg100_correct} and \ref{fig:eg100_incorrect} show the samples drawn from the EPIC-Kitchens-100 dataset. We can observe that the model can narrow to the specific verbs of activities based on the context. For example, in video \textbf{(a)}, predictions converge to verb \textit{"wash"} or \textit{"squeeze"}. In addition, video \textbf{(b)} recognized \textit{"milk"} in the first four frames and then switched to the (possible) consequence object \textit{"cereal"}. The video \textbf{(c)} contains minimal movements in the scene change of cooking activity; as a result, our model consistently adhered to relevant predictions but struggled to anticipate the object until the last moment. Moreover, video \textbf{(d)} shows a challenging example where the subject's intention is vague during the entire observation timespan.

Figure \ref{fig:eg100_incorrect} reveals some cases our model failed to anticipate. For example, in the video \textbf{(e)}, although the model grabs the correct verb, a wrong noun \textit{"pizza"} is consistently predicted. Other examples can be interpreted as wrong predictions due to a lack of visual observations (e.g., \textbf{(f)}), and due to too many possibilities to narrow the objects (e.g., \textbf{(g)}, \textbf{(h)}).
\newline

\begin{figure*}
    \centering
    \includegraphics[width=0.89\textwidth]{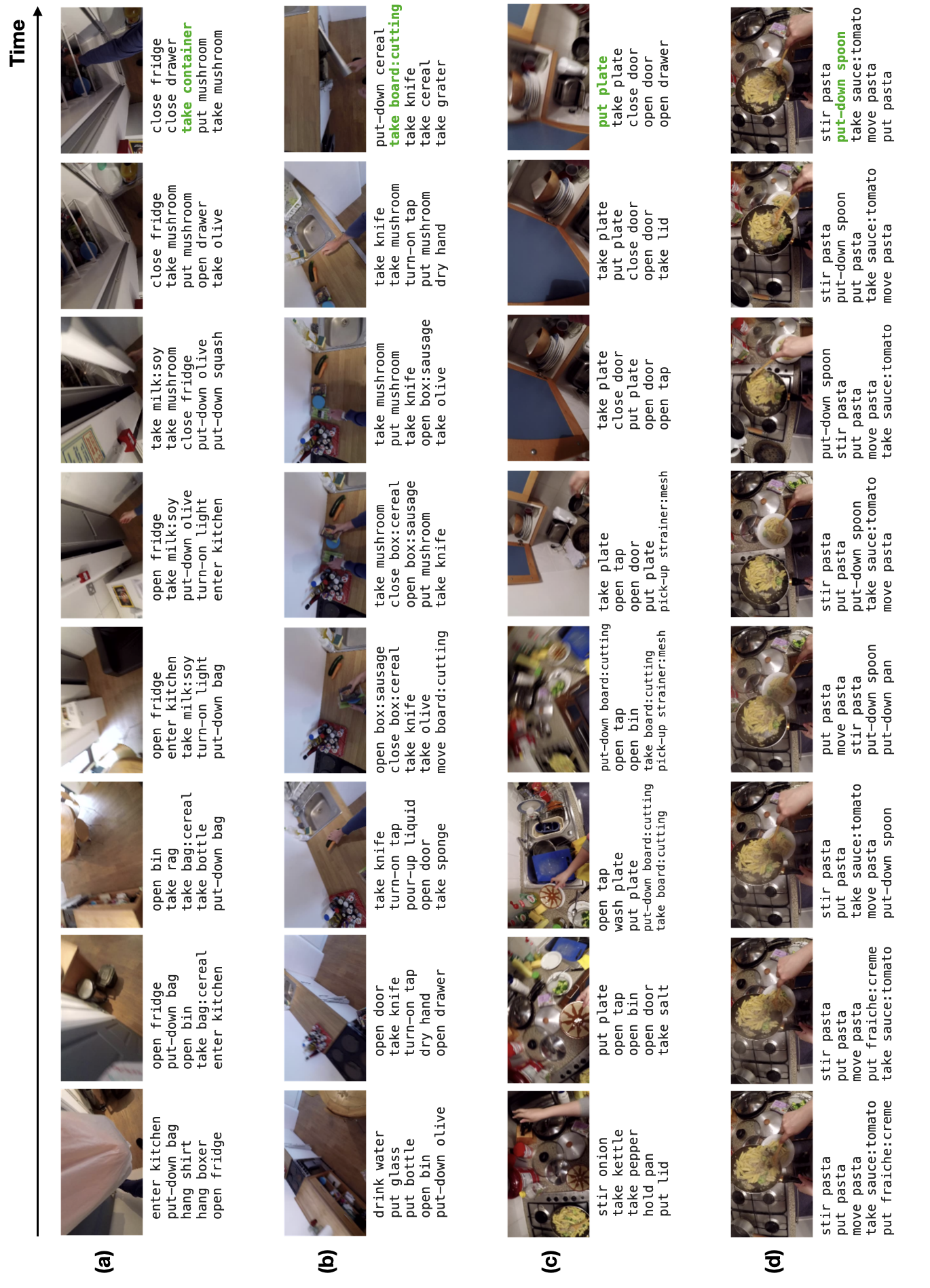}
    \caption{Some \textbf{correct} predictions illustrates on the \textbf{EPIC-Kitchens-55} validation set. The figure shows four video clips (the last eight frames presented) and the corresponding top-5 action anticipations ($\tau_a=1s$). The ground truth is highlighted in bold green.}
    \label{fig:eg55_correct}
\end{figure*}

\noindent \textbf{EPIC-Kitchens-55.}
Figure~\ref{fig:eg55_correct} shows successful cases with the last eight frames that the model observed. Note that the target noun object in videos \textbf{(a)} and \textbf{(b)} are invisible. However, the model can infer correctly based on past predictions propagated by inductive attention. Furthermore, video \textbf{(c)} requires the model to catch the object \textit{"plate"} that appears in the first three frames (it goes out of view in the rest of the frames); our model successfully captures the hint. Finally, video \textbf{(d)} holds the prediction results relevant to cooking pasta.

On the other hand, the sample \textbf{(e)} contains no hint about the anticipated object \textit{"tofu"} and predicts \textit{"container"} instead. Similarly, there are some mispredicted nouns found in videos \textbf{(f)} and \textbf{(h)}; and verbs in \textbf{(g)}.
\newline

\begin{figure*}
    \centering
    \includegraphics[width=0.88\textwidth]{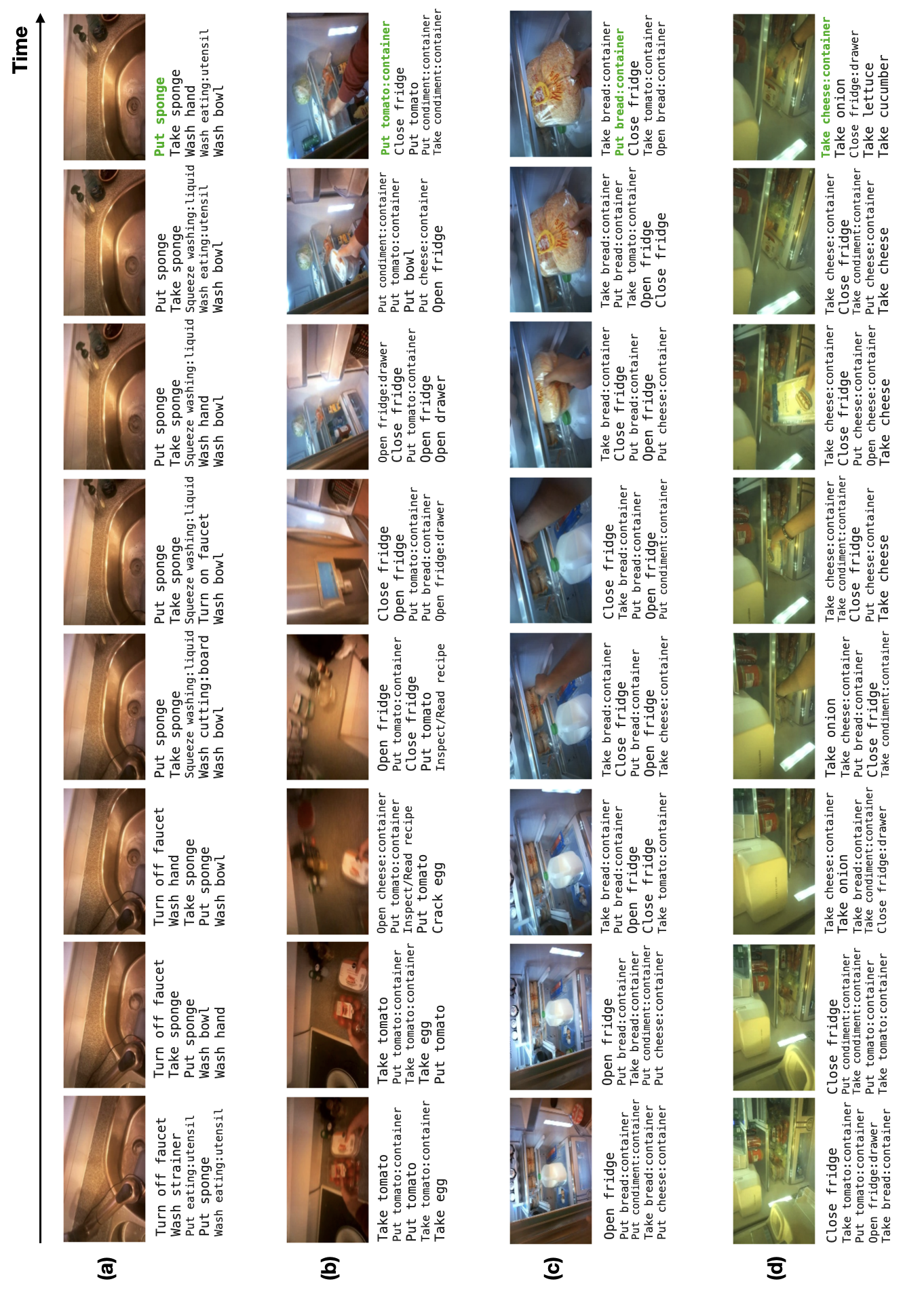}
    \caption{Some \textbf{correct} predictions illustrates on the \textbf{EGTEA Gaze+} validation set. The figure shows four video clips (the last eight frames presented) and the corresponding top-5 action anticipations ($\tau_a=0.5s$). The ground truth is highlighted in bold green.}
    \label{fig:egtea_gaze_correct}
\end{figure*}

\noindent \textbf{EGTEA Gaze+.}
We highlight the video \textbf{(b)} in Figure~\ref{fig:egtea_gaze_correct}, where the prediction can only be issued correctly by being aware of \textit{tomato:container} which only appears in the first two frames, that is six frames away from the last frame. Also, video \textbf{(c)} is with the subtle movement about the \textit{bread:container}. Note how it was taken out and then put back in the last three frames.

\begin{figure*}
    \centering
    \includegraphics[width=0.88\textwidth]{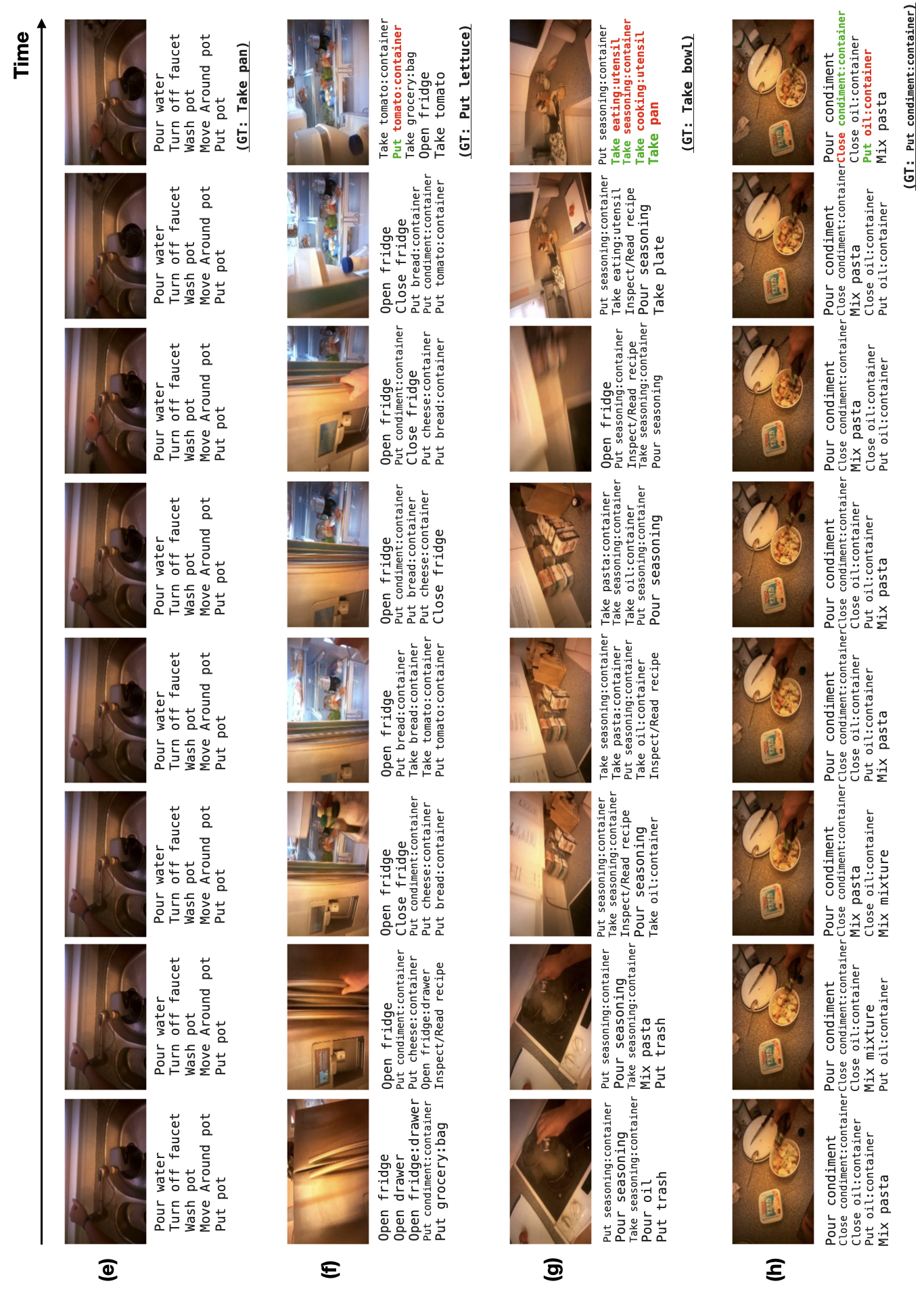}
    \caption{Some \textbf{incorrect} predictions illustrates on the \textbf{EGTEA Gaze+} validation set. The figure shows four video clips (the last eight frames presented) and the corresponding top-5 action anticipations ($\tau_a=0.5s$). The ground truth is revealed at the end. The correct verb and noun of action are highlighted in green.}
    \label{fig:egtea_gaze_incorrect}
\end{figure*}

Figure~\ref{fig:egtea_gaze_incorrect} reveals some incorrect predictions, primarily due to the insufficient evidence that can be found in the observations (e.g., \textbf{(e)} and \textbf{(f)}), or due to too many possibilities (e.g., \textbf{(g)}; however, note the confidence on verbs), or mixtures in \textbf{(h)}.

\begin{figure*}
    \centering
    \includegraphics[width=0.89\textwidth]{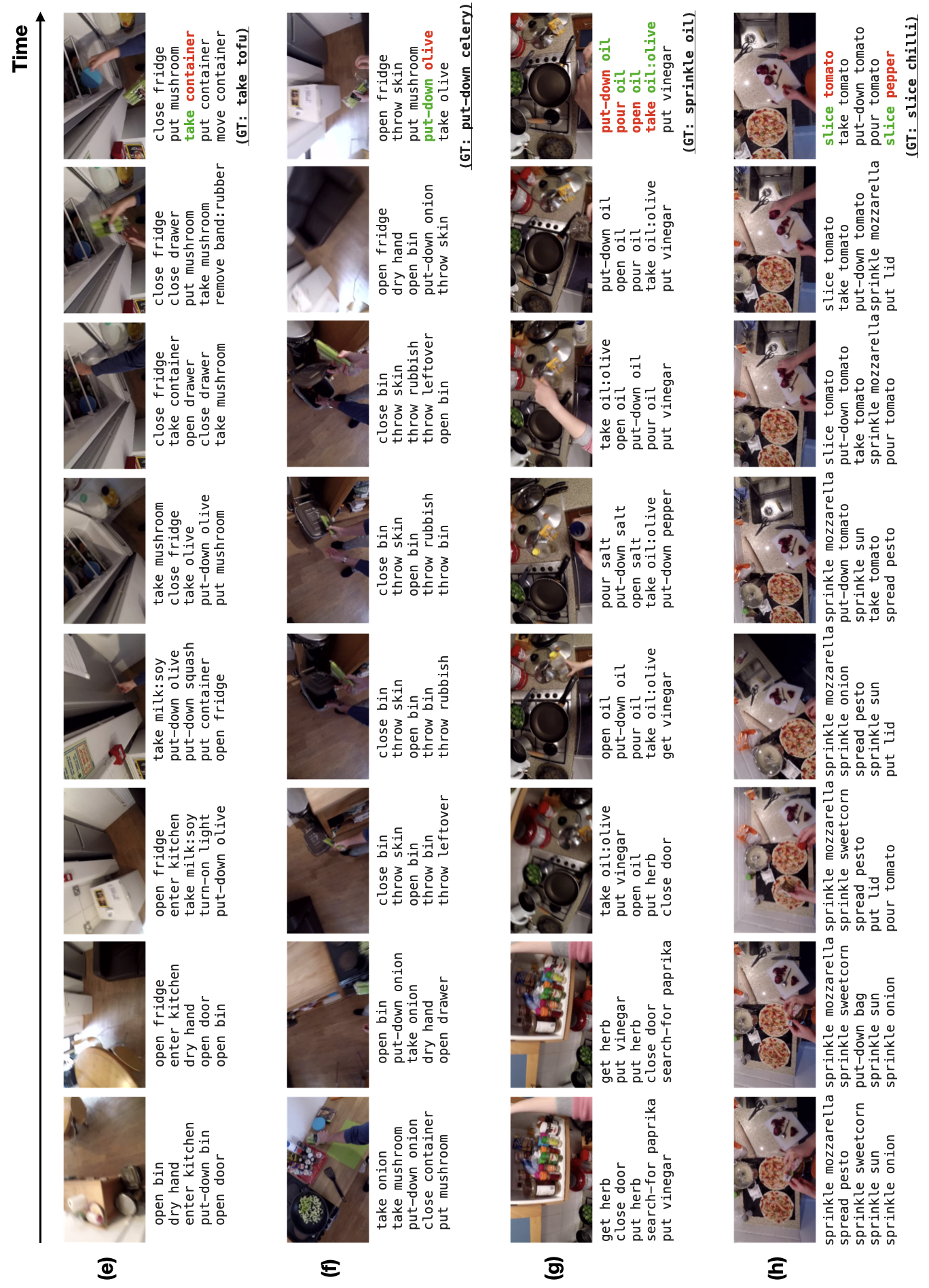}
    \caption{Some \textbf{incorrect} predictions illustrates on the \textbf{EPIC-Kitchens-55} validation set. The figure shows four video clips (the last eight frames presented) and the corresponding top-5 action anticipations ($\tau_a=1s$). The ground truth is revealed at the end. The correct verb and noun of action are highlighted in green.}
    \label{fig:eg55_incorrect}
\end{figure*}

% \end{document}

\end{document}